\documentclass[journal]{IEEEtran}

\usepackage{amsmath}
\usepackage{array}
\usepackage{url}
\usepackage{amsthm}
\usepackage{amssymb}
\usepackage{bm}
\usepackage{booktabs}
\usepackage{epstopdf}
\usepackage{epsfig}
\usepackage{multirow}
\usepackage{color}
\usepackage{graphicx}
\usepackage{cite}

\ifCLASSINFOpdf
\else
\fi
\begin{document}

\title{Embedded Deep Bilinear Interactive Information and Selective Fusion for Multi-view Learning}

\author{Jinglin Xu, Wenbin Li, Jiantao Shen, Xinwang Liu, Peicheng Zhou, Xiangsen Zhang,\\ Xiwen Yao, and Junwei Han
\thanks{Jinglin Xu, Jiantao Shen, Peicheng Zhou, Xiangsen Zhang, Xiwen Yao, and Junwei Han are with School of Automation, Northwestern Polytechnical University, China (e-mail: \{xujinglinlove, shenjt97, zpc19881119, xszhang116, yaoxiwen517, junweihan2010\}@gmail.com).}
\thanks{Wenbin Li is with the State Key Laboratory for Novel Software Technology, Nanjing University, China. (e-mail: liwenbin.nju@gmail.com)}
\thanks{Xinwang Liu is with National University of Defense Technology, China. (e-mail: xinwangliu@nudt.edu.cn)}
}


\maketitle

\begin{abstract}
As a concrete application of multi-view learning, multi-view classification improves the traditional classification methods significantly by integrating various views optimally. Although most of the previous efforts have been demonstrated the superiority of multi-view learning, it can be further improved by comprehensively embedding more powerful cross-view interactive information and a more reliable multi-view fusion strategy in intensive studies. To fulfill this goal, we propose a novel multi-view learning framework to make the multi-view classification better aimed at the above-mentioned two aspects. That is, we seamlessly embed various intra-view information, cross-view multi-dimension bilinear interactive information, and a new view ensemble mechanism into a unified framework to make a decision via the optimization. In particular, we train different deep neural networks to learn various intra-view representations, and then dynamically learn multi-dimension bilinear interactive information from different bilinear similarities via the bilinear function between views. After that, we adaptively fuse the representations of multiple views by flexibly tuning the parameters of the view-weight, which not only avoids the trivial solution of weight but also provides a new way to select a few discriminative views that are beneficial to make a decision for the multi-view classification. Extensive experiments on six publicly available datasets demonstrate the effectiveness of the proposed method.
\end{abstract}


\IEEEpeerreviewmaketitle

\section{Introduction}
\IEEEPARstart{C}{lassification} has been a fundamental technique of machine learning and broadly applied in image classification \cite{Cheng2017Remote,han2019pcnn,Xu2017Feature}, image reconstruction \cite{du2018reconstructing}, cross-modal retrieval \cite{yu2018category}, gaze tracking \cite{lian2018multiview}, visual tracking \cite{Zhang2015Object,cheng2019learning}, multi-view boosting \cite{peng2017multiview}, multi-view biological data \cite{feng2019supervised} and so on. The typical classification methods mainly include Logistic Regress, K-Nearest Neighbor, Decision Tree, and Support Vector Machines (SVM), where the kernel trick is an important innovation of SVM since it can not only utilize the convex optimization method with convergence to learn the nonlinear model but also make the implementation of kernel function (such as Radial Basis Function (RBF)) high-efficiency. Unfortunately, the computation cost of the kernel methods is too large for the large dataset. The deep learning method, therefore, is proposed to overcome the limitations of the kernel methods. For example, \cite{hinton2006a} has been demonstrated that the performance of Neural Networks is better than that of SVM with RBF kernel on the MNIST dataset.

Recently, with the rapid development of data mining techniques, in many scientific data analysis tasks, data are often collected through different ways, such as various sensors, as usually a single way cannot comprehensively describe the entire information of a data instance. In this case, different kinds of features of each data instance can be considered as different views of this instance, where each view may capture a specific facet of this instance. Importantly, the combination of multiple views can provide a more comprehensive view of this instance. Obviously, the multi-view features can be utilized to sufficiently represent an instance compared to the single view features. As an important branch of multi-view learning \cite{zhao2017multi}, the multi-view classification \cite{xu2019multiview,xie2019multiview,tang2017multiview} uses multiple distinct representations of data and models a multi-view learning framework to perform the classification task.

Canonical Correlation Analysis (CCA) \cite{Hotelling1936Relations} and Kernel CCA (KCCA) \cite{Bach2002Kernel,Hardoon2004Canonical,Sun2013A} show their abilities of effectively modeling the relationship between two or more views. However, they still have some limitations on capturing high-level associations between different views. To be specific, CCA ignores the nonlinearities in multi-view data and KCCA may suffer from the effect of small data when the data acquisition in one or more modalities is expensive or otherwise limited. Therefore, in the early efforts, \cite{Lai1999A} investigates a neural network implementation of CCA and maximizes the correlation between the output of the networks for each view. \cite{Hsieh2000Nonlinear} formulates a nonlinear CCA method using three feedforward neural networks, where the first network maximizes the correlation between two canonical variates, while the remaining two networks map from the canonical variates back to the original two sets of variables. However, the above CCA-based methods have been proposed for many years and are needed to be improved.

Inspired by the recent successes of deep neural networks \cite{SalakhutdinovH09,Yoshua2013Representation}, correlation can be naturally applied to multi-view neural network learning to learn deep and abstract multi-view interactive information. The deep neural networks extension of CCA (Deep CCA) \cite{Andrew2013Deep} learns representations of two views by using multiple stacked layers and maximizes the correlation of two representations. Later, \cite{WangALB15} proposes a deep canonically correlated auto-encoder (DCCAE) by combining the advantages of both Deep CCA and auto-encoder. Specifically, DCCAE consists of two auto-encoders and optimizes the combination of canonical correlation between two learned bottleneck representations and the reconstruction errors of the auto-encoders. Similar to the principle of DCCAE, \cite{Feng2014CRC} proposes a correspondence auto-encoder (Corr-AE) via constructing correlations between the hidden representations of two unimodal deep auto-encoders. \cite{Rastegar2016MDL} suggests exploiting the cross weights between the representations of views for gradually learning interactions of the modalities (views) in a multi-modal deep auto-encoder network. Theoretical analysis of \cite{Rastegar2016MDL} shows that considering these interactions from a high level provides more intra-modality (intra-view) information.

In addition, a number of multi-view analysis methods \cite{DietheHS10,Kan2012Multi} have been proposed. Based on Fisher Discriminant Analysis (FDA) \cite{Mika01animproved}, both regularized two-view FDA and its kernel extension can be cast as equivalent disciplined convex optimization problems. Then, \cite{DietheHS10} introduces Multi-view Fisher Discriminant Analysis (MFDA) that learns classifiers in multiple views, by minimizing the variance of the data along with the projection while maximizing the distance between the average outputs for classes over all the views. However, MFDA can only be used for binary classification problems. \cite{Kan2012Multi} proposes a Multi-view Discriminant Analysis (MvDA), which seeks for a discriminant common space by maximizing the between-class and minimizing the within-class variations, across all the views. Later, based on bilinear models \cite{Tenenbaum2014Separating} and general graph embedding framework \cite{Shuicheng2007Graph}, \cite{Sharma2012Generalized} introduces Generalized Multi-view Analysis (GMA). As an example of GMA, Generalized Multi-view Linear Discriminant Analysis (GMLDA) finds a set of projection directions in each view that tries to separate different contents\rq\ class means and unify different views of the same class in the common subspace.

Based on the above analyses, although CCA-based deep neural networks can learn the multi-view interactive information and DA-based methods can consider the discriminative information, there is not a unified framework that simultaneously embeds the intra-view information, cross-view interactive information, and a reliable multi-view fusion strategy.

To tackle this issue, we propose a novel multi-view learning framework named as MvNNBiIn, which integrates both the intra-view information and cross-view multi-dimension bilinear interactive information for each view and designs a multi-view selective loss fusion. Specifically, (a) multiple intra-view information coming from different faceted representations, ensures the diversity and complementarity among different views to enhance multi-view learning. (b) The cross-view multi-dimension bilinear interactive information dynamically learns multi-dimension bilinear interactive information from different bilinear similarities. Each bilinear similarity is calculated by the similarity between intra-view information via the bilinear function. Therefore, the multi-dimension bilinear interactive information comprehensively models the relationships between views by learning different metric matrices. (c) The multi-view selective loss fusion is to calculate multiple losses for multiple views and fuses them in an adaptive weighting way with the selective strategy. This selective strategy can choose several discriminative views that are beneficial to make a decision for the multi-view classification, by tuning the sparseness of the weight vector.

It is worth mentioning that we have developed a preliminary work \cite{xu2020deep} named \emph{deep embedded complementary and interactive information for multi-view classification} (denoted as MvNNcor). In this paper, our proposed MvNNBiIn method extends and improves the MvNNcor method significantly. Their major differences are summarized as follows. On one hand, MvNNcor utilizes the cross-correlations between attributes of multiple representations to generate interactive information. In contrast, our MvNNBiIn models a novel cross-view multi-dimension bilinear interactive information which consists of different bilinear similarities for each view with respect to another view. Each bilinear similarity is generated by the similarity between intra-view information passed by the bilinear function. On the other hand, MvNNcor uses the multi-view fusion strategy to integrate multiple views. By contrast, our MvNNBiIn designs a novel view ensemble mechanism to select more discriminative views that are beneficial to the multi-view classification.

We summarize the main contributions of the proposed MvNNBiIn method as follows.
\begin{itemize}
\item We propose a unified framework, which seamlessly embeds intra-view information, cross-view multi-dimension bilinear interactive information, and a novel view ensemble mechanism, to make a decision during the optimization and improve the classification performance.
\item We model the cross-view interactive information by capturing the multi-dimension bilinear interactive information which is calculated by simultaneously learning multiple metric matrices via the bilinear function between views. It comprehensively models the relationships between different views.
\item We develop a new view ensemble mechanism which not only selects some discriminative views but also fuses them via an adaptive weighting method. The selective strategy can ensure that the selected views are beneficial to the multi-view classification.
\item We perform extensive experiments on several publicly available datasets to demonstrate the effectiveness of our model.
\end{itemize}

The rest of this paper is organized as follows. The Preliminary Knowledge is introduced in Section \ref{preliminary_knowledge}. We formulate the framework of MvNNBiIn in Section \ref{proposed_method}. Section \ref{optimization} briefly provides a tractable and skillful optimization method of MvNNBiIn. Section \ref{experiments} evaluates MvNNBiIn on several public datasets, followed by some theoretical and empirical analyses of experiments. The conclusion and further works are shown in Section \ref{conclusion}.

\section{Preliminary Knowledge}\label{preliminary_knowledge}

In this section, we briefly review several multi-view learning methods from the perspectives of CCA-based methods and MvDA method.

\subsection{CCA-based Methods}

CCA \cite{Hotelling1936Relations} is popular for its capability of modeling the relationship between two or more sets of variables. CCA computes a shared embedding of both or more sets of variables through maximizing the correlations among the variables among these sets. CCA has been widely used in the multi-view learning tasks to generate low-dimensional representations \cite{Rasiwasia2010A,Liang2008A}. Improved generalization performance has been witnessed in areas including dimensionality reduction \cite{Avron2013Efficient}, clustering \cite{Blaschko2008Correlational,ChaudhuriKLS09}, regression \cite{Kakade2007Multi,Mcwilliams2013Correlated}, word embeddings \cite{Dhillon2011Multi,DhillonRFU12,Gong2014A}, and discriminant learning \cite{Tae2007Discriminative,Ya2012Discriminant}.

Supposing that $\bm{X}^v|_{v=1,2}\in\mathbb{R}^{d_v\times N}$ is a data matrix for the $v$-th view, CCA tends to find the linear projections $\bm{w}^v|_{v=1,2}\in\mathbb{R}^{d_v}$ which make the instances from two data matrices $\bm{X}^v|_{v=1,2}$ maximally correlated in the projected space. Therefore, CCA is modeled as the following constrained optimization problem,
\begin{equation}
\begin{split}
  &\underset{\bm{w}^v|_{v=1,2}}{\max}\bm{w}^{1\top}\bm{X}^1\bm{X}^{2\top}\bm{w}^2\\
  &s.t.\ \bm{w}^{v\top}\bm{X}^v\bm{X}^{v\top}\bm{w}^v=1,v=1,2
\end{split}
\end{equation}
where $\bm{X}^v|_{v=1,2}$ are centralized.

KCCA is an kernel extension of CCA for pursuing maximally correlated nonlinear projections. Let $\bm{K}_v$ denote the kernel matrix such that $\bm{K}_v=\bm{H}\tilde{\bm{K}_v}\bm{H}$, where $[\tilde{\bm{K}_v}]_{ij}=k_v(\bm{x}_i^v,\bm{x}_j^v)=<\!\phi_v(\bm{x}_i^v),\phi_v(\bm{x}_j^v)\!>$, $\phi_v:\mathcal{X}\!\rightarrow\!\mathcal{H}$, $k_v:\mathcal{X}\times\mathcal{X}\rightarrow\mathbb{R}$. $\bm{H}=\bm{I}-\frac{1}{n}\bm{1}\bm{1}^\top$ is a centering matrix and $\bm{1}\in\mathbb{R}^n$ denotes a vector of all ones. Therefore, KCCA is formulated as the following optimization problem,
\begin{equation}
\begin{split}
  &\underset{\bm{\alpha},\bm{\beta}}{\max}\frac{\bm{\alpha}^\top\bm{K}_1\bm{K}_2\bm{\beta}}
  {\sqrt{\bm{\alpha}^\top\bm{K}_1^2\bm{\alpha}\bm{\beta}^\top\bm{K}_2^2\bm{\beta}}}\\
  &s.t.\ \bm{\alpha}^\top\bm{K}_1^2\bm{\alpha}=1,\bm{\beta}^\top\bm{K}_2^2\bm{\beta}=1
\end{split}
\end{equation}
where $\bm{\alpha}$ and $\bm{\beta}$ are the coefficient vectors optimized by KCCA.

As a DNN extension of CCA, DCCA utilizes two DNNs $f_1$ and $f_2$ to extract nonlinear features for each view, and then maximizes the canonical correlation between $f_1(\bm{X}^1)$ and $f_2(\bm{X}^2)$. That is,
\begin{equation}
\begin{split}
  &\underset{\bm{W}_{f_1},\bm{W}_{f_2},\bm{U},\bm{V}}{\max}\frac{1}{N}Tr\left(\bm{U}^\top f_1(\bm{X}^1)f_2(\bm{X}^2)^\top\bm{V}\right)\\
s.t.
&\bm{U}^\top\left(\frac{1}{N}f_1(\bm{X}^1)f_1(\bm{X}^1)^\top+r_1\bm{I}\right)\bm{U}=\bm{I}\\
&\bm{V}^\top\left(\frac{1}{N}f_2(\bm{X}^2)f_2(\bm{X}^2)^\top+r_2\bm{I}\right)\bm{V}=\bm{I}\\
&\bm{u}_i^\top f_1(\bm{X}^1)f_2(\bm{X}^2)^\top\bm{v}_j=0,\ \text{for}\ i\neq j
\end{split}
\end{equation}
where $\bm{U}\!=\![\bm{u}_1,\!\cdots\!,\bm{u}_d]\!\in\!\mathbb{R}^{d_1\times d}$ and $\bm{V}\!=\![\bm{v}_1,\!\cdots\!,\bm{v}_d]\!\in\!\mathbb{R}^{d_2\times d}$ are the CCA directions that project the DNN outputs, and $(r_1,r_2)>0$ are the regularization parameters for the sample covariance estimation.

Inspired by both CCA and reconstruction-based objectives, DCCAE constructs a model by consisting of two auto-encoders and optimizing the combination of canonical correlation between the learned representations and the reconstruction errors of the auto-encoders.
That is,
\begin{equation}
\begin{split}
&\underset{\bm{U},\bm{V},\bm{W}_{f_1},\atop\bm{W}_{f_2},\bm{W}_{g_1},\bm{W}_{g_2}}{\min}-\frac{1}{N}Tr\left(\bm{U}^\top f_1(\bm{X}^1)f_2(\bm{X}^2)^\top\bm{V}\right)\\
&+\frac{r}{N}\sum_{i=1}^N(\|\bm{x}^1_i\!-\!g_1(f_1(\bm{x}^1_i))\|^2\!+\!\|\bm{x}^2_i\!-\!g_2(f_2(\bm{x}^2_i))\|^2)
 \\
s.t.
&\bm{U}^\top\left(\frac{1}{N}f_1(\bm{X}^1)f_1(\bm{X}^1)^\top+r_1\bm{I}\right)\bm{U}=\bm{I}\\
&\bm{V}^\top\left(\frac{1}{N}f_2(\bm{X}^2)f_2(\bm{X}^2)^\top+r_2\bm{I}\right)\bm{V}=\bm{I}\\
&\bm{u}_i^\top f_1(\bm{X}^1)f_2(\bm{X}^2)^\top\bm{v}_j=0,\ \text{for}\ i\neq j
\end{split}
\end{equation}
where $g_1$ and $g_2$ are the reconstruction networks for each view, and $r$ is the trade-off parameter.

\subsection{MvDA Method}

MvDA attempts to find $M$ linear transforms $\bm{w}^v|_{v=1}^M$ that project the samples from $M$ views to one discriminant common space, respectively, where the between-class variation is maximized while the within-class variation is minimized. Defined $\bm{X}^v=\{\bm{x}^v_{ik}\}|_{k,i=1}^{C,n^v_k}$ as the samples from the $v$-th view, where $\bm{x}^v_{ik}\in\mathbb{R}^{d_v}$ is the $i$-th sample from the $v$-th view of the $k$-th class of $d_v$ dimension and $C$ denotes the number of classes, and $n^v_k$ is the number of samples from the $v$-th view of the $k$-th class.

The samples from $M$ views are then projected to the same common space by using $M$ view-specific linear transforms. The projected results are denoted as
$\bm{Y}=\{\bm{y}^v_{ik}=\bm{w}^{v\top}\bm{x}^v_{ik}\}|_{v,k,i=1}^{M,C,n^v_k}$. In the common space, according to our goal, the between-class variation $\bm{S}_B^y$ from all views should be maximized while the within-class variation $\bm{S}_W^y$ from all views should be minimized. Therefore, the objective is formulated as a generalized Rayleigh quotient,
\begin{equation}
\begin{split}
(\bm{w}^{1\star},\cdots,\bm{w}^{M\star})=\underset{\bm{w}^1,\cdots,\bm{w}^M}{\arg\max}\frac{Tr(\bm{S}_B^y)}{Tr(\bm{S}_W^y)}
\end{split}
\end{equation}
where the within-class scatter matrix $\bm{S}_W^y$ and the between-class scatter matrix $\bm{S}_B^y$ are computed as below,
\begin{equation}
\begin{split}
&\bm{S}_W^y=\sum_{k=1}^C\sum_{v=1}^M\sum_{i=1}^{n^v_k}(\bm{y}^v_{ik}-\bm{\mu}_k)(\bm{y}^v_{ik}-\bm{\mu}_k)^\top\\
&\bm{S}_B^y=\sum_{k=1}^Cn_k(\bm{\mu}_k-\bm{\mu})(\bm{\mu}_k-\bm{\mu})^\top
\end{split}
\end{equation}
where $n_k=\sum_{v=1}^Mn^v_k$ denotes the number of samples of the $k$-th class in all views, $N=\sum_{k=1}^Cn_k$ is the number of samples from all the classes and all the views, $\bm{\mu}_k$ is the mean of all the samples of the $k$-th class over all the views in the common space, $\bm{\mu}$ is the mean of all the samples over all views in the common space. That is,
\begin{equation}
\begin{split}
\bm{\mu}_k=\frac{1}{n_k}\sum_{v=1}^M\sum_{i=1}^{n^v_k}\bm{y}^v_{ik},\
\bm{\mu}=\frac{1}{N}\sum_{k=1}^C\sum_{v=1}^M\sum_{i=1}^{n^v_k}\bm{y}^v_{ik}
\end{split}
\end{equation}
\begin{figure*}[ht]
\begin{center}
\includegraphics[width=\linewidth]{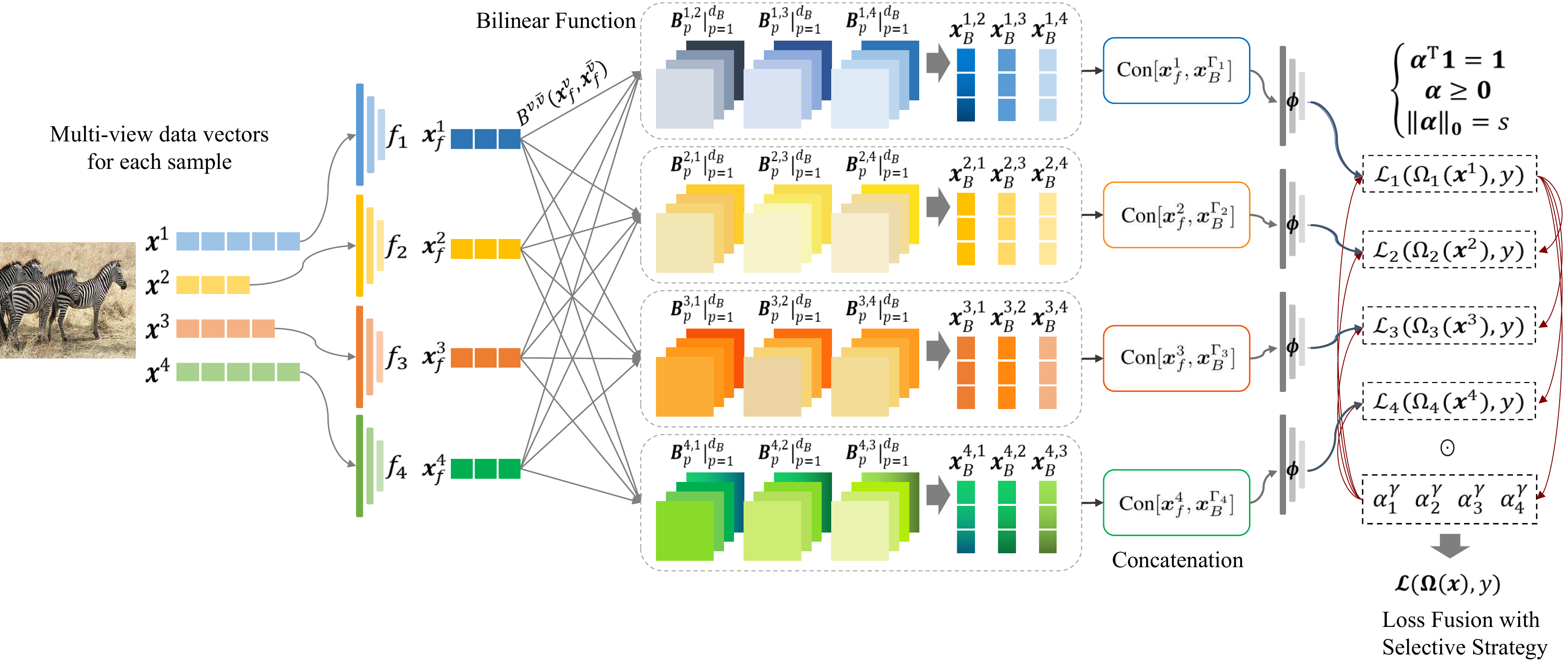}
\end{center}
\caption{The architecture of our proposed method MvNNBiIn. It consists of four parts: $M$ intra-view extracted networks $\{f_v\}_{v=1}^M$, cross-view bilinear interactive modules, the combination module of intra-view and cross-view information, and the multi-view selective loss fusion strategy. }
\label{flowchart}
\end{figure*}

\section{The Proposed Method}\label{proposed_method}

\noindent In this section, we propose the architecture of our model MvNNBiIn, depicted in Figure \ref{flowchart}, which consists of four parts: various intra-view extracted networks, multi-dimension bilinear interactive modules between views, combination of intra-view and cross-view information, and the multi-view selective loss fusion strategy.

\subsection{Various Intra-view Information Extraction}

Given an instance $\bm{x}$, we utilize $M$ views $\{\bm{x}^v\}_{v=1}^M$ to denote its $M$ kinds of visual features, which ensures the diverse and complementary information during the multi-view learning. Defining a set of neural networks $\{f_v\}_{v=1}^M$, each $f_v$ projects $\bm{x}^v$ from $\mathbb{R}^{d_v}$ into $\mathbb{R}^d$, which captures the high-level intra-view information for the $v$-th view, that is,
\begin{equation}
\bm{x}_f^v=f_v(\bm{x}^v),
\end{equation}
where $\bm{x}_f^v\in\mathbb{R}^d$ and $f_v$ is a neural network with $L_f$ layers,
\begin{equation}
\bm{h}_{f_v}^l=\bm{\sigma}(\bm{W}_{f_v}^l\bm{h}_{f_v}^{l-1}+\bm{b}_{f_v}^l).
\end{equation}
where $\bm{W}_{f_v}^l\!\in\!\mathbb{R}^{m_l\times m_{l-1}}$ denotes the weight matrix and $\bm{b}_{f_v}^l\!\in\!\mathbb{R}^{m_l}$ denotes the bias vector, and $\bm{h}_{f_v}^l\!\in\!\mathbb{R}^{m_l}$ is the output of the $l$-th layer, and $\bm{\sigma}$ is the activation function applied component-wise. Specifically, $l\!=\!1,\cdots,L_f$, $m_0\!=\!d_v$, $m_{L_f}\!=\!d$, $\bm{h}_{f_v}^0\!=\!\bm{x}^v\!$, and $\bm{h}_{f_v}^{L_f}\!=\!\bm{x}_f^v$.
It is worth mentioning that $\{f_v\}_{v=1}^M$ are trained coordinatively by solving the optimization problem shown in subsection \ref{loss_function}, where the parameters of $f_v$ are not shared with those of $\{f_{\bar{v}}|\bar{v}=\{1,\cdots,M\}\setminus v\}$. Although we utilize a multi-layer perceptron as $f_v$, it also can be replaced with any deterministic neural networks with the input and output layers.

\subsection{Multi-dimension Bilinear Interactive Information}

\begin{figure}[t]
\begin{center}
\includegraphics[width=0.6\linewidth]{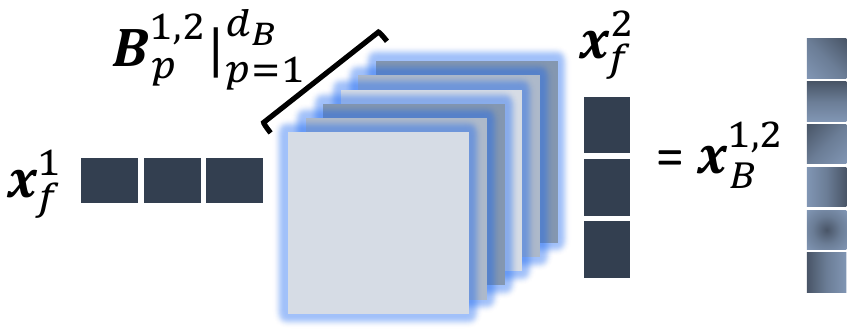}
\end{center}
\caption{The construction of the multi-dimension bilinear interactive information of each pair of views. Taking the first view $\bm{x}_f^1$ as an example, we calculate its multi-dimension bilinear interactive information relative to the second view $\bm{x}_f^2$ by learning multiple metric matrics, i.e., $(v,\tilde{v})=(1,2)$, and then obtain $\bm{x}_B^{1,2}\in\mathbb{R}^{d_B}$.}
\label{bilinear-details}
\end{figure}

Inspired by the metric learning, the multi-dimension bilinear interactive information is proposed to construct the second-order feature interactions on various intra-view information. To be specific, given the $v$-th view $\bm{x}_f^v$ and the $\bar{v}$-th view $\bm{x}_f^{\bar{v}}$, the multi-dimension bilinear interactive information between different views can be formulated via the bilinear function $B^{v,\bar{v}}$, that is,
\begin{equation}
\bm{x}_B^{v,\bar{v}}=B^{v,\bar{v}}(\bm{x}_f^v,\bm{x}_f^{\bar{v}})
                    =\begin{bmatrix}
                    \bm{x}_f^{v\top}\bm{B}^{v,\bar{v}}_1\bm{x}_f^{\bar{v}}+b^{v,\bar{v}}_1\\
                    \vdots\\
                    \bm{x}_f^{v\top}\bm{B}^{v,\bar{v}}_{d_B}\bm{x}_f^{\bar{v}}+b^{v,\bar{v}}_{d_B}
\end{bmatrix}
\end{equation}
where $\bm{x}_B^{v,\bar{v}}\in\mathbb{R}^{d_B}$, $\bm{B}^{v,\bar{v}}_p\in\mathbb{R}^{d\times d}$, $p=1,\cdots,d_B$, and $\bm{b}_p^{v,\bar{v}}\in\mathbb{R}$ is the bias. There are multiple metric matrices (i.e., $\bm{B}^{v,\bar{v}}_1,\cdots,\bm{B}^{v,\bar{v}}_{d_B}$) to be learned simultaneously. Intuitively, Figure \ref{bilinear-details} shows the construction of the multi-dimension bilinear interactive information of each pair of views.

For the $v$-th view, we define $\Gamma_v$ as a set contained $M-1$ view pairs relative to the $v$-th view, that is,
\begin{equation}
\Gamma_v=\{(v,\bar{v})|\bar{v}=\{1,\cdots,M\}\setminus v\},
\end{equation}
where $v\in\{1,\cdots,M\}$ and each view pair $(v,\bar{v})$ is undirected. For the $v$-th view, the bilinear interactive information of the $v$-th view relative to other $M-1$ views is collected into a set $\bm{x}_B^{\Gamma_v}$,
\begin{equation}
  \bm{x}_B^{\Gamma_v}=\{\bm{x}_B^{v,\bar{v}}|(v,\bar{v})=\Gamma_v\}
\end{equation}
where the $v$-th set $\Gamma_v$ contains $M-1$ view pairs for the $v$-th view and there are $M$ sets $\Gamma_v|_{v=1}^M$ for $M$ views.

\subsection{Combination and Prediction of Each View}

According to the above two subsections, we combine the intra-view information $\bm{x}_f^v$ and multi-dimension bilinear interactive information $\bm{x}_B^{\Gamma_v}$ of the $v$-th view as follows,
\begin{equation}
\bm{x}^{\Gamma_v}=\text{Con}[\bm{x}_f^v,\bm{x}_B^{\Gamma_v}]
\end{equation}
where Con denotes the concatenation operation.

After that, $\bm{x}^{\Gamma_v}\!\in\!\mathbb{R}^{d+d_B}$ is passed through $\phi$ to obtain the prediction $\bm{z}^v$ of the $v$-th view, that is,
\begin{equation}
\begin{split}
&\bm{z}^v=\phi(\bm{x}^{\Gamma_v})
\end{split}
\end{equation}
where $C$ is the number of classes and $\bm{z}^v\!\in\!\mathbb{R}^C$ produces a distribution over the possible classes for each view. $\phi$ is a neural network with $L_\phi$ layers, that is,
\begin{equation}
\begin{split}
\bm{h}_{\phi}^l=\bm{\sigma}(\bm{W}_\phi^l\bm{h}_\phi^{l-1}+\bm{b}_\phi^l)
\end{split}
\end{equation}
where $\bm{W}_\phi^l\!\in\!\mathbb{R}^{n_l\times n_{l-1}}$ denotes the weight matrix and $\bm{b}_\phi^l\!\in\!\mathbb{R}^{n_l}$ denotes the bias vector, and $\bm{h}_\phi^l\!\in\!\mathbb{R}^{n_l}$ is the output of the $l$-th layer, and $\bm{\sigma}$ is the activation function applied component-wise. In particular, $l\!=\!1,\cdots,L_\phi$, $n_0\!=\!d+d_B$, $n_{L_\phi}\!=\!C$, $\bm{h}_\phi^0\!=\!\bm{x}^{\Gamma_v}$, and $\bm{h}_\phi^{L_\phi}\!=\!\bm{z}^v$. Here, the parameters of $\phi$ are shared among $M$ views.

\subsection{Multi-view Selective Loss Fusion Strategy}\label{loss_function}

In this subsection, we design a novel view ensemble mechanism, which is to calculate multiple losses for multiple views and fuses them in an adaptive weighting way with the selective strategy. This selective strategy can choose several discriminative views that are beneficial to the multi-view classification, by tuning the sparseness of the weight vector. That is described as,
\begin{equation}\label{weight_optimization}
\begin{split}
&\underset{\bm{\alpha}}{\min}\sum_{v=1}^M\alpha_v^{\gamma}\mathcal{L}^v(\bm{z}^v,y)\\ &s.t.\ \bm{\alpha}^\top\bm{1}\!=\!\bm{1},\bm{\alpha}\!\geq\!\bm{0},\|\bm{\alpha}\|_0=s
\end{split}\end{equation}
where
\begin{align}
\mathcal{L}^v(\bm{z}^v,y)=-\log\left(\frac{\exp(z^v_y)}{\sum_{o=1}^C\exp(z_o^v)}\right),
\end{align}
where $\bm{\alpha}\in\mathbb{R}^M$ is the weight for each view, $y\in\mathbb{R}$ denotes the common label information of an instance for all the views, and $\mathcal{L}^v(\bm{z}^v,y)$ is the cross-entropy loss of the $v$-th view. $\gamma>1$ is the power exponent parameter of the weight $\alpha_v$ for the $v$-th view, which controls the weight distribution of different views flexibly and avoids the trivial solution of $\bm{\alpha}$ during the classification. $\|\bm{\alpha}\|_0=s$ is used to constrain the sparseness of the weight vector $\bm{\alpha}$, where $s\in\mathbb{N}_+$ denotes the number of nonzero elements in $\bm{\alpha}$. Crucially, the $L_0$-norm constraint is able to capture the global relationship among different views and to achieve the view-wise sparsity, which realizes selecting a few discriminative views can improve the performance during the multi-view classification.
Intuitively, the architecture of our proposed method MvNNBiIn is shown in Figure \ref{flowchart}.

\section{Optimization}\label{optimization}

We utilize the alternate optimization method to update the parameters $\{\{f_v\}_{v=1}^M, B, \phi\}$ and the view-weight $\bm{\alpha}$, respectively. For convenience, $B$ denotes the bilinear function set $\{B^{v,\bar{v}}|(v,\bar{v})\!=\!\Gamma_v\}_{v=1}^M$ and $\psi$ is used to denote the concatenation operation in Figure \ref{gradient}.

\subsection{Update $\{f_v\}_{v=1}^M$, $B$, and $\phi$}

\begin{figure}[t]
\begin{center}
\includegraphics[width=0.9\linewidth]{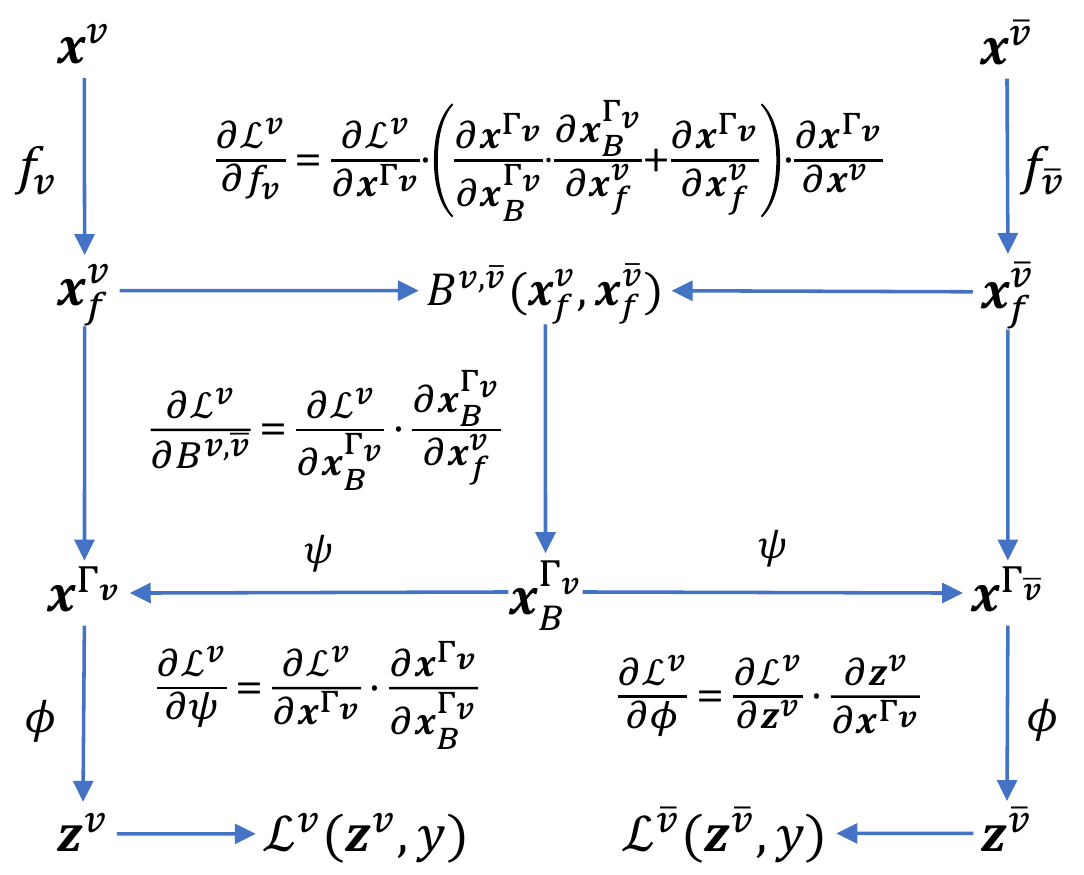}
\end{center}
\caption{The gradient computations of our proposed MvNNBiIn method during the multi-view classification.}
\label{gradient}
\end{figure}

We fix $\bm{\alpha}$ and update $\{f_v\}_{v=1}^M$, $B$, and $\phi$, which utilizes Adam with batch normalization and the autograd package in PyTorch to train. Figure \ref{gradient} briefly shows the gradient computations of our proposed MvNNBiIn method.
\begin{table*}[ht]
\renewcommand\arraystretch{1.2}
\centering
\caption{The descriptions of six publicly available datasets, i.e., Caltech101, Caltech20, AWA, NUSOBJ, Reuters, and SUN. Some abbreviations are defined as follows, Fea: public Feature, Dim: Dimensionality, WM: Wavelet Moments, CENT: CENTRIST, CH: Color Histogram, LSS: Local Self-Similarity, PHOG: Pyramid HOG, RGSIFT: Color SIFT, CM: block-wise Color Moments, CORR: color Correlogram, EDH: Edge Direction Histogram, WT: Wavelet Texture, GEOMAP: Geometric Map.}
\begin{tabular}{ccccccccccccc}
\toprule
\multirow{2}{*}{Views} &
\multicolumn{2}{c}{Caltech101} &
\multicolumn{2}{c}{Caltech20} &
\multicolumn{2}{c}{AWA} &
\multicolumn{2}{c}{NUSOBJ} &
\multicolumn{2}{c}{Reuters} &
\multicolumn{2}{c}{SUN} \\
& Fea & Dim & Fea & Dim & Fea & Dim & Fea & Dim & Fea & Dim & Fea & Dim\\
\midrule
1  & Gabor & 48  & Gabor & 48  & CH    & 2688 & CH   & 64  &  English & 21531 & GIST   & 256\\
2  & WM    & 40  & WM    & 40  & LSS   & 2000 & CM   & 225 &  French  & 24892 & GEOMAP & 512\\
3  & CENT  & 254 & CENT  & 254 & PHOG  & 252  & CORR & 144 &  German  & 34251 & TEXTON & 512\\
4  & HOG   & 1984& HOG   & 1984& SIFT  & 2000 & EDH  & 73  &  Italian & 15506 & /      & /\\
5  & GIST  & 512 & GIST  & 512 & RGSIFT& 2000 & WT   & 128 &  Spanish & 11547 & /      & /\\
6  & LBP   & 928 & LBP   & 928 & SURF  & 2000 & /    &/    &  /       & /     & /      & /\\
\midrule
\# Samples & \multicolumn{2}{c}{9914} & \multicolumn{2}{c}{2386} & \multicolumn{2}{c}{30475} & \multicolumn{2}{c}{30000} & \multicolumn{2}{c}{18758} & \multicolumn{2}{c}{99250}\\
\# Classes & \multicolumn{2}{c}{102}  & \multicolumn{2}{c}{20}   & \multicolumn{2}{c}{50}    & \multicolumn{2}{c}{31}    & \multicolumn{2}{c}{6}     & \multicolumn{2}{c}{397}\\
\bottomrule
\end{tabular}
\label{datasets}
\end{table*}

\subsection{Update $\bm{\alpha}$}

We learn the non-negative normalized weight $\alpha_v$ for each view and assign the higher weight to more discriminative view. Therefore, we fix the parameters $\{\{f_v\}_{v=1}^M,B,\phi\}$ and update $\bm{\alpha}$ by solving the optimization problem (\ref{weight_optimization}).

To efficiently minimize the problem (\ref{weight_optimization}), we define a function $\varphi$ on $\bm{\mathcal{L}}(\bm{z},y)=[\mathcal{L}^1(\bm{z}^1,y),\cdots,\mathcal{L}^M(\bm{z}^M,y)]\in\mathbb{R}^M$, that is,
\begin{equation}\label{permutation}
  \varphi(\bm{\mathcal{L}}(\bm{z},y))=\bm{\mathcal{L}}(\bm{z},y)\bm{P}
\end{equation}
where $\bm{P}$ is a permutation matrix which results in permuting elements of $\bm{\mathcal{L}}(\bm{z},y)$ along the ascending order, i.e., $\mathcal{L}^{p(1)}(\bm{z}^{p(1)},y)\leq\cdots\leq\mathcal{L}^{p(M)}(\bm{z}^{p(M)},y)$. Based on equation (\ref{permutation}), we select the first $s$ smallest elements and optimize their corresponding weights $\alpha_{p(v)}|_{v=1}^s$, meanwhile, setting the rest $M-s$ weights $\alpha_{p(v)}|_{v=s+1}^M$ as zeros. Therefore, the problem (\ref{weight_optimization}) is equivalent to the following problem by absorbing the $L_0$-norm constraint $\|\bm{\alpha}\|_0=s$ into the objective function,
\begin{equation}\label{weight_optimization_transform}
\begin{split}
  &\underset{\alpha_{p(v)}|_{v=1}^s}{\min}\sum_{v=1}^s\alpha_{p(v)}^\gamma\mathcal{L}^{p(v)}(\bm{z}^{p(v)},y)\\
  &s.t.\alpha_{p(v)}\geq0,\ \sum_{v=1}^s\alpha_{p(v)}=1
  \end{split}
\end{equation}

Through the Lagrangian Multiplier method, the Lagrangian function of problem (\ref{weight_optimization_transform}) is:
\begin{equation}\label{lagrangian}
  L(\alpha_{p(v)},\lambda)=\sum_{v=1}^s\alpha_{p(v)}^\gamma\mathcal{L}^{p(v)}(\bm{z}^{p(v)},y)
  +\lambda(\sum_{v=1}^s\alpha_{p(v)}-1)
\end{equation}
where $\lambda$ is the Lagrangian multiplier. Taking the derivatives of $L(\alpha_{p(v)},\lambda)$ with respect to $\alpha_{p(v)}$ and $\lambda$, respectively, and setting them to zeros, there is,
\begin{equation}\label{alpha_s}
  \alpha_{p(v)}=\frac{\mathcal{L}^{p(v)}(\bm{z}^{p(v)},y)^{\frac{1}{1-\gamma}}}
  {\sum_{w=1}^s\mathcal{L}^{p(w)}(\bm{z}^{p(w)},y)^{\frac{1}{1-\gamma}}}
\end{equation}
where $v=1,\cdots,s$ and $s$ is the sparsity of $\bm{\alpha}$.

To sum up, the optimal solution of problem (\ref{weight_optimization}) can be calculated by,
\begin{equation}\label{alpha}
  \alpha_{p(v)}=\left\{\begin{matrix}
\frac{\mathcal{L}^{p(v)}(\bm{z}^{p(v)},y)^{\frac{1}{1-\gamma}}}
  {\sum_{w=1}^s\mathcal{L}^{p(w)}(\bm{z}^{p(w)},y)^{\frac{1}{1-\gamma}}},&v\!=\!1,\cdots,s\\
0,&v\!=\!s\!+\!1,\cdots,M
\end{matrix}\right.
\end{equation}

Therefore, the scheme for MvNNBiIn can be summarized as follows. Firstly, the $\{f_v\}_{v=1}^M$ are used to learn the complementary and diverse intra-view information from multi-view representations. Secondly, the intra-view information is used for training the bilinear function set $B$, which captures the cross-view interactive information. Thirdly, the intra-view and cross-view interactive information for each view is integrated by a concatenation operation, and then the concatenation is fed into $\phi$ to calculate the cross-entropy loss. Finally, the trained $\{f_v\}_{v=1}^M$, $B$ and $\phi$ can be optimized by solving the problem (\ref{weight_optimization}) and the weight distribution can be obtained, which contributes to infer the label during the multi-view classification.

\section{Experiments}\label{experiments}

In this section, we evaluate the performance of our proposed MvNNBiIn  method on six publicly available datasets.

\subsection{Datasets}

We follow many state-of-the-art multi-view methods, which treat different kinds of pre-extracted feature vectors for an image as different views to make the experimental comparisons fair. Different kinds of pre-extracted features present different facets of the images, such as color, texture, shape, and geographic information, yet the above information is not intuitive. We describe all the publicly available datasets in TABLE \ref{datasets}.

Specifically, Caltech101 \cite{fei2007learning,li2015large} and Caltech20 \cite{li2015large} datasets consist of images of objects belonging to 102 (101 classes plus one background clutter class) and 20 classes, respectively, and each dataset is described as 6 views.
AWA (Animals with Attributes) \cite{lampert2009learning} dataset provides 6 kinds of features (6 views) for attribute base classification.
NUSOBJ dataset is a subset of NUS-WIDE \cite{chua2009nus}, which describes each object image using 5 types of low-level features (5 views).
Reuters \cite{amini2009learning} dataset is used for document categorization and written in 5 different languages (5 views).
SUN is a subset of SUN397 (Scene Categorization Benchmark) \cite{Xiao2010SUN,Xiao2016SUN} and utilizes 3 kinds of public features matrices (3 views) to represent each image.

Referring to the previous works \cite{Andrew2013Deep,wang2015deep}, we split each dataset into three parts, that is, 70\% samples for training, 20\% samples for validating, and 10\% samples for testing. We utilize the classification accuracy (e.g., Top@1 accuracy and Top@5 accuracy) to evaluate the performance of all the methods and report the final results in TABLEs \ref{dB_size}, \ref{compare_with_others}, and \ref{ablation}.

\begin{figure*}[t]
\begin{center}
\includegraphics[width=\linewidth]{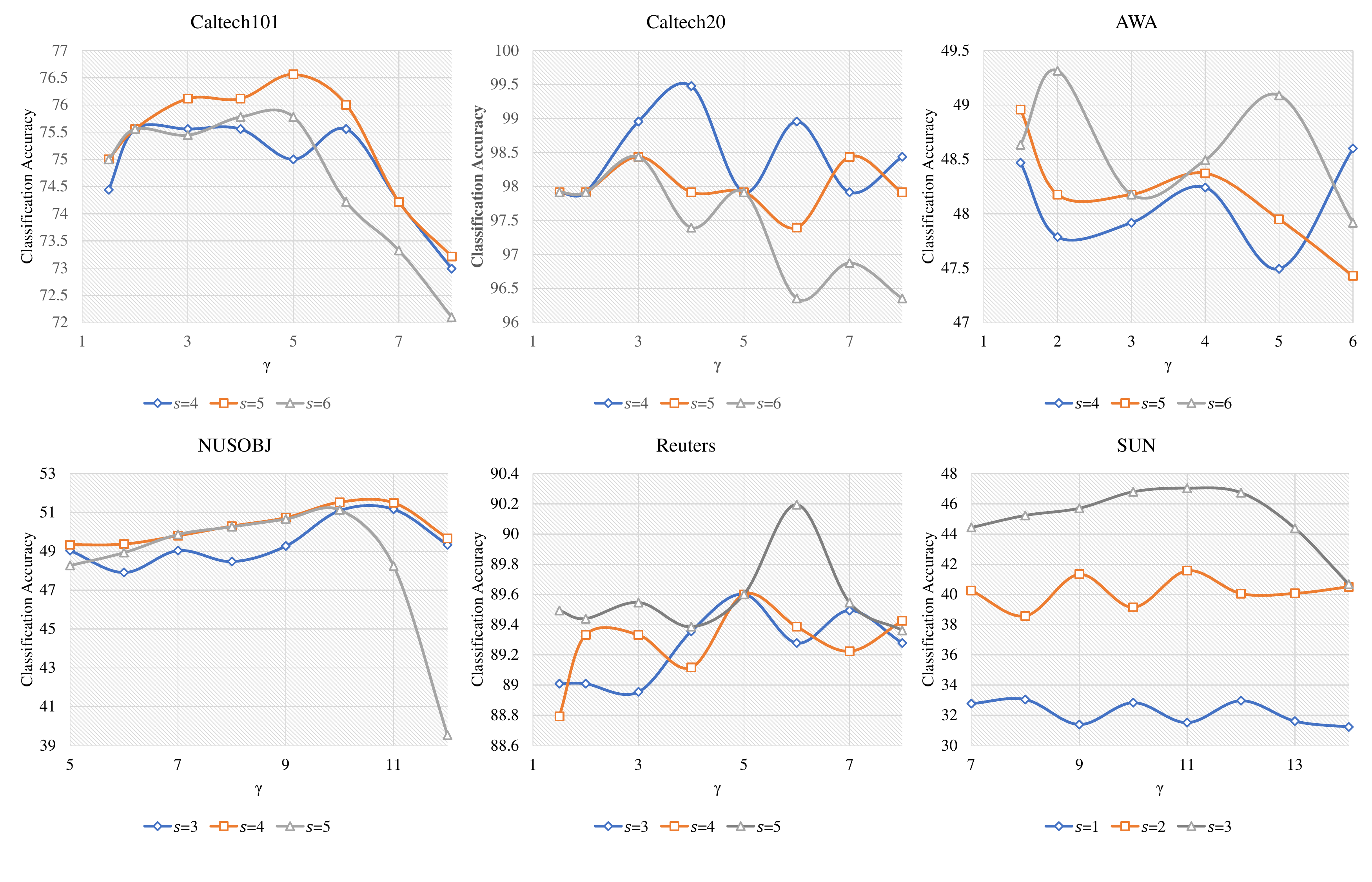}
\end{center}
\caption{The effect of sparseness on various views in our proposed method MvNNBiIn. $s$ denotes the number of nonzero elements in learned weights $\bm{\alpha}$. Since different datasets contain different numbers of views, we explore $s$ from $\{4,5,6\}$ on Caltech101, Caltech20, and AWA datasets; from $\{3,4,5\}$ on NUSOBJ and Reuters datasets; $\{1,2,3\}$ on SUN dataset. $\gamma$ is the power exponent parameter of the weight $\alpha_v$ and we investigate $\gamma$ in the range of $[1.5,8]$ on Caltech101, Caltech20, and Reuters datasets; in the range of $[1.5,6]$ on AWA dataset; in the range of $[5,12]$ on NUSOBJ dataset; in the range of $[7,14]$ on SUN dataset.}
\label{gamma_s_visualization}
\end{figure*}
\subsection{Experimental Settings}

\subsubsection{Comparison Methods}
We first compare our MvNNBiIn method with several state-of-the-art multi-view methods, including SVMcon, DCCA \cite{Andrew2013Deep}, DCCAE \cite{wang2015deep}, DeepLDA \cite{DorferKW15}, MvDA \cite{Kan2015Multi}, MvDN \cite{kan2016multideep}, and MvNNcor \cite{xu2020deep}, in Table \ref{compare_with_others}.
In particular, SVMcon is a baseline that concatenates all the views and feeds into the SVM classifier. DCCA and DCCAE belong to the CCA-based method inputting two views. DeepLDA, MvDA, and MvDN are the  Discriminant Analysis based method, where DeepLDA inputs the concatenation of all the views and feeds into a deep neural network followed by LDA; MvDA inputs multiple views with the same dimensionality; MvDN is the non-linear version of LDA, which uses deep neural networks to replace of the linear transformations.

What\rq s more, Table \ref{ablation} demonstrates the effectiveness of three important parts of our proposed MvNNBiIn, i.e., intra-view information, cross-view bilinear interactive information, and a novel view ensemble mechanism, respectively. The highest performance is obtained when all the parts are available while the performance is lower when any part is absent.
\begin{table*}[ht]
\renewcommand\arraystretch{1.2}
\centering
\caption{The investigation of the dimensionality $d_B$ of the bilinear function $B^{v,\bar{v}}$ on different datasets. $d_B$ is set as $50$, $100$, $200$, and $400$, respectively, which reflects the combination proportion of intra-view information and cross-view bilinear interactive information for each view. Due to CUDA out of memory, the results of $d_B=400$ cannot be obtained except for NUSOBJ dataset.}
\begin{tabular}{ccccccccccccc}
\toprule
\multirow{2}{*}{$d_B$} &
\multicolumn{2}{c}{Caltech101} &
\multicolumn{2}{c}{Caltech20} &
\multicolumn{2}{c}{AWA} &
\multicolumn{2}{c}{NUSOBJ} &
\multicolumn{2}{c}{Reuters} &
\multicolumn{2}{c}{SUN} \\
& Top@1 & Top@5 & Top@1 & Top@5 & Top@1 & Top@5 & Top@1 & Top@5 & Top@1 & Top@5 & Top@1 & Top@5\\
\midrule
50  & 76.228 & 87.277 & 97.397 & 98.958 & 48.210 & 74.977 & 51.529 & 84.142 & 89.116 & 99.838 & 43.770 & 71.069\\
100 & 75.781 & 88.058 & 97.397 & 98.958 & 47.428 & 74.740 & 51.529 & 84.774 & 89.224 & 99.731 & 46.200 & 72.742\\
\textbf{200} & \textbf{76.562} & \textbf{89.063} & \textbf{98.958} & \textbf{99.473} & \textbf{49.316} & \textbf{75.521} & \textbf{51.529} & \textbf{84.843} & \textbf{90.194} & \textbf{99.838} & \textbf{47.036} & \textbf{73.458}\\
400 & / & / & / & / & / & / & 51.496 & 84.441 & / & / & / & /\\
\bottomrule
\end{tabular}
\label{dB_size}
\end{table*}
\begin{table*}[t]
\renewcommand{\arraystretch}{1.2}
\setlength\tabcolsep{16pt}
	\centering
    \caption{Comparison results of MvNNBiIn and several state-of-the-art methods on all the datasets.}
    \label{compare_with_others}
\begin{tabular}{lcccccc}
\toprule
Method   &  Caltech101 & Caltech20 & AWA & NUSOBJ & Reuters & SUN\\
\midrule
SVMcon  & 47.901 & 83.827 & 31.044 & 42.719 & 88.180 & 38.200\\
DeepLDA     & 45.649 & 76.508 & 25.598 & 20.320 & 84.907 & /\\
MvDA      & 45.200 & 76.276 & 9.788 & 11.457 & 78.831 & /\\
DCCA      & 66.159 & 86.504 & 20.677 & 28.753 & 64.917 & 16.116\\
DCCAE    & 26.894 & 50.267 & 13.484 & 27.477 & 56.530 & /\\
MvDN     & 70.214 & 94.833 & 42.359 & 47.241 & 88.339 & 40.538\\
MvNNcor  & 76.002 & 97.924 & 47.687 & 52.049 & 89.276 & 45.632 \\
MvNNBiIn & \textbf{76.562} & \textbf{98.958} & \textbf{49.316} & \textbf{51.529} & \textbf{90.194} & \textbf{47.036}\\
\bottomrule
\end{tabular}
\end{table*}
\begin{table*}[t]
\renewcommand{\arraystretch}{1.2}
\setlength\tabcolsep{11pt}
	\centering
    \caption{Ablation experiments of our proposed MvNNBiIn on all the datasets.}
    \label{ablation}
\begin{tabular}{lcccccccc}
\toprule
$\phi$ & \checkmark & \checkmark & \checkmark & \checkmark & \checkmark & \checkmark & \checkmark & \checkmark\\
$\{f_v\}_{v=1}^M$ & & \checkmark & & \checkmark & \checkmark & \checkmark & \checkmark & \checkmark\\
$B$ & & & \checkmark & \checkmark & & & \checkmark & \checkmark\\
$\{\alpha_v\}_{v=1}^M$ & & & & & \checkmark & & \checkmark & \\
$\{\alpha_v\}_{v=1}^M,\|\bm{\alpha}\|_0=s$ & & & & & & \checkmark & & \checkmark\\
\midrule
Caltech101 & 60.044 & 73.237 & 49.330 & 72.210 & 73.750 & 74.038 & 75.781 & \textbf{76.562}\\
Caltech20 & 94.792 & 96.096 & 82.292 & 97.917 & 96.615 & 97.917 & 98.438 & \textbf{98.958}\\
AWA & 37.341 & 43.424 & 36.003 & 41.862 & 45.427 & 46.953 & 48.893 & \textbf{49.316}\\
NUSOBJ & 48.847 & 50.366 & 36.469 & 47.739 & 48.055 & 50.673 & 51.130 & \textbf{51.529}\\
Reuters & 88.578 & 88.766 & 88.739 & 88.793 & 89.170 & 89.536 & 89.978 & \textbf{90.194}\\
SUN & 20.605 & 39.798 & 31.341 & 3.942 & 30.655 & 40.936 & 46.794 & \textbf{47.036}\\
\bottomrule
\end{tabular}
\end{table*}

\subsubsection{Parameters Setup}
Deep LDA is a fully connected neural network consisted of three hidden layers, i.e., 400, 200, and 300 units equipped with ReLU activation function. In DCCAE, there is a feature extraction network and a reconstruction network for each view, where each network is a fully connected network consisted of three hidden layers, i.e., 400, 200, and 300 units equipped with ReLU activation function, followed by a linear output of $C$ (for feature extraction network) /$d_v$ (for reconstruction network) units. The capacities of the above networks are the same as those of their counterparts in DCCA and MvDN.

In our MvNNBiIn, two kinds of networks $\{\{f_v\}_{v=1}^M, \phi\}$ and the bilinear function set $B\!=\!\{B^{v,\bar{v}}|(v,\bar{v})\!=\!\Gamma_v\}_{v=1}^M$ are needed to learn. Each $f_v$ is a fully connected network consisted of two hidden layers (i.e., 400 and 200 units equipped with ReLU activation function). $\phi$ consists of 200$\cdot M$ input units and 300 hidden units equipped with ReLU activation function, followed by a linear output of $C$ units. Each $B^{v,\bar{v}}$ is a bilinear function which outputs $d_B\!=\!200$ bilinear interactive units from each pair $(\bm{x}_f^v,\bm{x}_f^{\bar{v}})$.
That is, the inputs of $B^{v,\bar{v}}$ are the outputs of $f_v$ and $f_{\bar{v}}$, and the input of $\phi$ is concatenated by the outputs of $f_v$ and $\{B^{v,\bar{v}}|(v,\bar{v})=\Gamma_v\}$. The capacities of the above networks are the same as those of their counterparts in the ablation studies shown in Table \ref{ablation}. In MvNNcor, three kinds of networks $\{\{f_v\}_{v=1}^M, \psi, \phi\}$ are needed to learn, where $\{\{f_v\}_{v=1}^M, \phi\}$ are the same as the counterparts of MvNNBiIn, and $\psi$ consists of 200$^2$ input units and 200 hidden units with ReLU activation function.

In this paper, all the networks are optimized by Adam with batch normalization, where the learning rate is 10$^{-3}$, $\beta_1=0.5$, $\beta_2=0.9$, and batch size is 64. In addition, we vary $\gamma$ and $s$, respectively, to explore the influence of different values of $\gamma$ and $s$ on the classification accuracy. Based on the optimal $\gamma$ and $s$, we can learn the optimal model to achieve the highest classification accuracy. The results are shown in Figure \ref{gamma_s_visualization} where Caltech101, Caltech20, AWA, NUSOBJ, Reuters, and SUN datasets can achieve the best performance when $(\gamma,s)$ are set as $(5,5)$, $(4,4)$, $(2,6)$, $(10,4)$, $(6,5)$, and $(11,3)$, respectively.

Besides, to explore the proper combination proportion of the intra-view information and multi-dimension bilinear interactive information for each view, we investigate the dimensionality $d_B$ of the output of the bilinear function $B^{v,\bar{v}}$, i.e., setting $d_B$ as 50, 100, 200, and 400, respectively, and the results are reported in Table \ref{dB_size}. It can be seen that the classification accuracy is the highest when $d_B=200$, which shows that the combination proportion also has an impact on the classification performance.
\begin{figure*}[t]
\begin{center}
\includegraphics[width=0.8\linewidth]{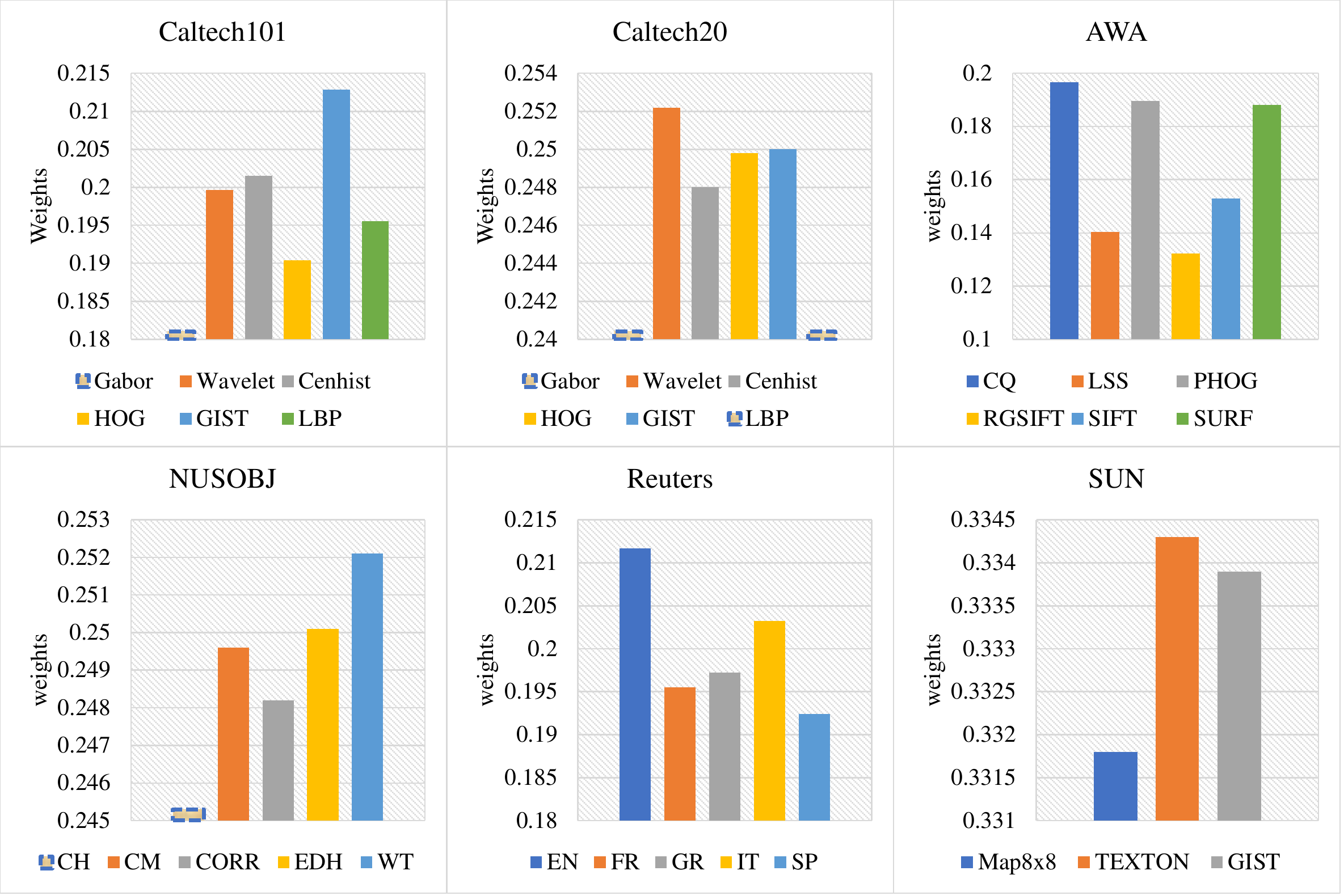}
\end{center}
\caption{The learned weights for different views through MvNNBiIn for Caltech101/20, AWA, NUSOBJ, Reuters, and SUN datasets. The blue dashed boxes respectively in Caltech101, Caltech20, and NUSOBJ datasets denote that the view weights are zeros.}
\label{weights}
\end{figure*}

\subsection{Experimental Results}
Tables \ref{compare_with_others} and \ref{ablation} show the classification performance of all the methods, where Table \ref{compare_with_others} reports the experimental results of several recent methods and Table \ref{ablation} provides the results of ablation experiments.

Firstly, compared with the single-view methods, i.e., SVMcon and DeepLDA, our MvNNBiIn consistently outperforms them on all the datasets. For example, MvNNBiIn achieves 28.661\% and 30.913\% improvements, respectively, on the Caltech101 dataset. That is because the concatenation of all the views may reduce the interpretability of different views and ignore the cross-view interactive information during the multi-view classification. We compare our MvNNBiIn with MvDA and the classification accuracy of our MvNNBiIn is better than that of MvDA on all the datasets. For instance, our MvNNBiIn obtains 31.362\% improvements on the Caltech101 dataset, since the linear transformations of MvDA cannot deal well with some subtle but important structures in some challenging scenarios. Compared to MvDN, our MvNNBiIn achieves 6.348\% improvements on the Caltech101 dataset due to embedding the multi-dimension bilinear interactive information between different views.

Secondly, we compare our MvNNBiIn with the CCA-based methods, i.e., DCCA and DCCAE, and the results are reported in Table \ref{compare_with_others}. It can be seen that our MvNNBiIn performs better than DCCA and DCCAE, since these two methods are limited to the double-view input and unable to capture more diverse and complementary information from more views. For example, compared to DCCA and DCCAE, our MvNNBiIn achieves 28.639\% and 35.832\% improvements on the AWA dataset, respectively.

Thirdly, as the extension work of MvNNcor, our MvNNBiIn almost achieves better performance on all the datasets. For example, compared with MvNNcor on the AWA dataset, MvNNBiIn achieves 1.629\% improvements. For one reason, the cross-view multi-dimension bilinear interactive information of our MvNNBiIn is more able to capture the interactive information between different views. For another reason, our MvNNBiIn designs a novel view ensemble mechanism which can select more discriminative views and is more beneficial to the multi-view classification.

Besides, in the ablation studies of our MvNNBiIn shown in Table \ref{ablation}, taking the NUSOBJ dataset as an example, we can achieve 2.682\%, 1.163\%, 15.060\%, 3.790\%, 3.474\%, 0.856\%, and 0.399\% improvements compared with our framework respectively containing some of the following modules, i.e., $\phi$, $\phi\!+\!\{f_v\}_{v=1}^M$, $\phi\!+\!B$, $\phi\!+\!\{f_v\}_{v=1}^M\!+\!B$, $\phi\!+\!\{f_v\}_{v=1}^M\!+\!\{\alpha_v\}_{v=1}^M$, $\phi\!+\!\{f_v\}_{v=1}^M\!+\!\{\alpha_v\}_{v=1}^M\!+\!\|\bm{\alpha}\|_0\!=\!s$, and $\phi+\{f_v\}_{v=1}^M\!+\!B\!+\!\{\alpha_v\}_{v=1}^M$. These results successively demonstrate the effectiveness of integrating the intra-view information and multi-dimension bilinear interactive information between views as well as the adaptive weighting multi-view loss fusion with the selective strategy.

Moreover, Figure \ref{weights} shows the view-weights of each dataset learned by our MvNNBiIn, where the $x$-axis  means the indices of different views and the $y$-axis denotes the weight of each view. The higher weight indicates that the view provides more valuable information and makes more contribution.

\subsection{Discussion}
\begin{table*}[t]
\renewcommand{\arraystretch}{1.2}
\setlength\tabcolsep{12pt}
	\centering
	\caption{Comparison results of our MvNNBiIn and several deep convolutional neural network architectures on image datasets Caltech101, NUSOBJ, and SUN397.}
	\label{deep_features_results}
\begin{tabular}{lcccccc}
\toprule
\multirow{2}{*}{Methods}&
\multicolumn{2}{c}{Caltech101}&
\multicolumn{2}{c}{NUSOBJ}&
\multicolumn{2}{c}{SUN397}\\
& Pre-trained & Fine-tuned & Pre-trained & Fine-tuned & Pre-trained & Fine-tuned \\
\midrule
AlexNet & 85.860 & 87.830 &  59.240 & 58.980 & 41.250 & 42.590\\
GoogLeNet & 88.050 & 89.690 &  63.620 & 64.990 & 46.840 & 47.960\\
ResNet-101 & 90.240 & 92.430 &  69.690 & 70.260 & 55.320 & 55.600\\
VGGNet-16 & 86.180 & 90.240 &  64.720 & 67.570 & 48.190 & 50.380\\
MvNNBiIn  & \textbf{93.752} & \textbf{95.091} & \textbf{70.513} & \textbf{71.384} & \textbf{56.404} & \textbf{56.632} \\
\bottomrule
\end{tabular}
\end{table*}
\begin{table}[t]
\renewcommand{\arraystretch}{1.2}
\setlength\tabcolsep{8pt}
	\centering
    \caption{The experimental settings of all the networks on Caltech101 dataset.}
    \label{deep_features_settings_cal101}
\begin{tabular}{lcccc}
\toprule
Method   & Iterations & Batch size & Learning rate & \\
\midrule
AlexNet    & 50000 & 256& 0.0001 & SGD\\
GoogLeNet  & 50000 & 32 & 0.001  & SGD\\
ResNet-101 & 50000 & 8  & 0.00001& SGD\\
VGGNet-16  & 50000 & 50 & 0.0001 & SGD\\
\bottomrule
\end{tabular}
\end{table}
\begin{table}[t]
\renewcommand{\arraystretch}{1.2}
\setlength\tabcolsep{8pt}
	\centering
    \caption{The experimental settings of all the networks on NUSOBJ dataset.}
    \label{deep_features_settings_nus}
\begin{tabular}{lcccc}
\toprule
Method   & Iterations & Batch size & Learning rate & \\
\midrule
AlexNet    & 15000 & 256& 0.00001& SGD\\
GoogLeNet  & 30000 & 64 & 0.00001& SGD\\
ResNet-101 & 14000 & 8  & 0.00001& SGD\\
VGGNet-16  & 15000 & 50 & 0.00001& SGD\\
\bottomrule
\end{tabular}
\end{table}
\begin{table}[t]
\renewcommand{\arraystretch}{1.2}
\setlength\tabcolsep{8pt}
	\centering
    \caption{The experimental settings of all the networks on SUN397 dataset.}
    \label{deep_features_settings_sun}
\begin{tabular}{lcccc}
\toprule
Method   & Iterations & Batch size & Learning rate & \\
\midrule
AlexNet    & 50000 & 256& 0.00001& SGD\\
GoogLeNet  & 50000 & 64 & 0.00001& SGD\\
ResNet-101 & 50000 & 8  & 0.00001& SGD\\
VGGNet-16  & 50000 & 32 & 0.00001& SGD\\
\bottomrule
\end{tabular}
\end{table}

Actually, our MvNNBiIn is a generic framework that can improve the multi-view classification using not only the handcrafted features (such as HOG, LBP, or SURF) but also the deep model-learned features.

We apply four popular CNNs (including AlexNet \cite{krizhevsky2012imagenet}, GoogLeNet \cite{szegedy2015going}, VGGNet-16 \cite{SimonyanZ14a}, and ResNet-101 \cite{he2016deep}) on three publicly available image datasets (i.e., Caltech101, NUSOBJ, and SUN397), respectively, to generate the CNN feature representations including four transferred CNN features and four fine-tuned CNN features. Then, we compare our proposed MvNNBiIn method with the single-view CNN feature-based methods to demonstrate the superiority of our multi-view learning framework.

To be specific, for the transferred CNN feature-based methods, four off-the-shelf CNN models including VGGNet-16, ResNet-101, AlexNet, and GoogLeNet are first adopted as general feature extractors to extract CNN features and then linear one-versus-all SVMs ($C$=0.001) are used for classification. For the fine-tuned CNN feature-based methods, we fine-tune the aforementioned four CNN models on the training datasets to extract better CNN features and then adopt linear one-versus-all SVMs ($C$=0.001) for classification. For our proposed MvNNBiIn method, we regard four transferred CNN features as four views for each image and apply them into MvNNBiIn to perform the multi-view classification. Similarly, our proposed MvNNBiIn method is also performed on four fine-tuned CNN features.

The experimental results are shown in table \ref{deep_features_results} and the experimental settings of the fine-tuned CNN Models are shown in Tables \ref{deep_features_settings_cal101}$\sim$\ref{deep_features_settings_sun}. It can be seen that our proposed MvNNBiIn method outperforms the single-view CNN feature-based methods, respectively, on both transferred and fine-tuned CNN features, and averagely achieves 7.577\% (AlexNet), 5.551\% (GoogLeNet), 3.087\% (ResNet-101), and 6.212\% (VGGNet-16) improvements on the Caltech101 dataset. These results demonstrate the superiority of our proposed multi-view learning framework.

\section{Conclusion}\label{conclusion}

In this paper, we propose a novel multi-view learning framework denoted as MvNNBiIn which seamlessly embeds various intra-view information and cross-view multi-dimension bilinear interactive information as well as introducing a new view ensemble mechanism to jointly make decisions during the multi-view classification. Extensive experiments on several publicly available datasets demonstrate the effectiveness of our proposed MvNNBiIn method. Furthermore, we demonstrate the superiority of multi-view learning using the CNN feature representations, which provides a novel idea of fusing outputs of different deterministic neural networks in further work.

\bibliographystyle{IEEEtran}
\bibliography{IEEEabrv,jinglin}

\ifCLASSOPTIONcaptionsoff
  \newpage
\fi

\end{document}